\documentclass{article}

\usepackage[a4paper]{geometry}  %
\usepackage{graphicx}  %
\usepackage{caption}  %
\usepackage{subcaption}  %
\usepackage{amsmath, amssymb}  %
\usepackage[backend=bibtex, defernumbers=true, citestyle=numeric-comp, bibstyle=numeric, style=ieee]{biblatex}
\usepackage[rightcaption]{sidecap} %
\usepackage{fancyhdr}  %
\usepackage{multirow}  %
\usepackage{eso-pic}
\usepackage[hidelinks]{hyperref}  %

\AddToShipoutPictureBG{%
	\AtPageLowerLeft{%
		\put(50,50){\rotatebox{90}{\large\textcolor{gray}{Preprint. Final version published in: Bildverarbeitung für die Medizin 2025, Springer. DOI: \url{https://doi.org/10.1007/978-3-658-47422-5_71}.}}}%
	}%
}

\renewenvironment{abstract}{%
  \noindent\bfseries\abstractname:\normalfont}{}

\newcommand{\titlerunning}[1]{\fancyhead[L]{#1}}  %
\newcommand{\authorrunning}[1]{\fancyhead[R]{#1}}  %

\title{\textbf{Exploring the Integration of Key-Value Attention Into Pure and Hybrid Transformers for Semantic Segmentation}\footnote{This is a preprint of the following paper: Hwa, DeShin; Holmes, Tobias; Drechsler, Klaus, "Integration of Key-value Attention into Pure and Hybrid Transformers for Semantic Segmentation", published in Bildverarbeitung f\"ur die Medizin 2025, edited by Palm, Christoph; Breininger, Katharina; Deserno, Thomas; Handels, Heinz; Maier, Andreas; Maier-Hein, Klaus H.; Tolxdorff, Thomas M., 2025, Springer Fachmedien Wiesbaden. The final authenticated version is available online at: \url{https://doi.org/10.1007/978-3-658-47422-5_71}.}}
\author{
    DeShin Hwa,
    Tobias Holmes,
    Klaus Drechsler
    }
\date{} %

\titlerunning{Key-Value Attention in Vision Transformers}
\authorrunning{Hwa, Holmes \& Drechsler}  %

\begin{document}

\maketitle  %

\begin{abstract}
	While CNNs were long considered state of the art for image processing, the introduction of Transformer architectures has challenged this position. While achieving excellent results in image classification and segmentation, Transformers remain inherently reliant on large training datasets and remain computationally expensive. A newly introduced Transformer derivative named KV Transformer shows promising results in synthetic, NLP, and image classification tasks, while reducing complexity and memory usage. This is especially conducive to use cases where local inference is required, such as medical screening applications. We endeavoured to further evaluate the merit of KV Transformers on semantic segmentation tasks, specifically in the domain of medical imaging. By directly comparing traditional and KV variants of the same base architectures, we provide further insight into the practical tradeoffs of reduced model complexity. We ob- serve a notable reduction in parameter count and multiply accumulate operations, while achieving similar performance from most of the KV variant models when directly compared to their QKV implementation.
\end{abstract}

\section{Introduction}
\subsection{Motivation}
Convolutional Neural Networks (CNNs) have long been the state of the art for image classification and segmentation, particularly in medical imaging. However, CNNs have limitations in capturing long-range dependencies and global context due to their reliance on local receptive fields and convolutional kernels. This can lead to a loss of global information. Vision Transformers (ViTs) \cite{3628-dosovitskiy_16x16_words}, derived from Transformers \cite{3628-vaswani_attention_2017} used in NLP, address these issues with self-attention mechanisms that capture long-range dependencies and spatial relationships more effectively.
However, Transformers and ViTs remain more computationally expensive than traditional CNNs due to the quadratic complexity of the self-attention mechanism and the absence of weight-sharing, unlike CNNs, which use convolutional kernels.
To this end, \citeauthor{3628-borji_kv_transformer} \cite{3628-borji_kv_transformer} proposed the KV Transformer to investigate the possibility of reducing model complexity while retaining performance by omitting one of the vectors used in the attention mechanism. This is especially beneficial in resource-constrained settings like medical screenings, where efficient inference is preferable. Thus, we endeavour to assess the performance of KV attention in pure and hybrid Transformer models and for medical image segmentation tasks.
\subsection{Related work}\label{3628-sec:related-work}
\begin{figure}[hbtp]
	\centering
	\begin{subfigure}{0.49\textwidth}
		\includegraphics[width=\textwidth]{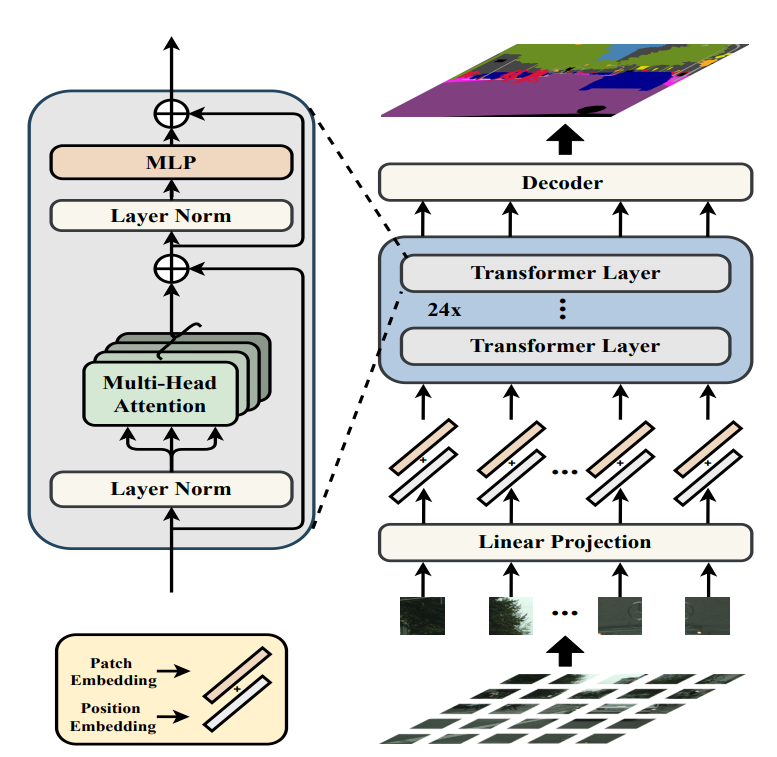}
		\caption{The standard SETR architecture \cite{3628-zheng_SETR}.}
		\label{3628-fig:setr-encoder}
	\end{subfigure}
	\hfill
	\begin{subfigure}{0.49\textwidth}
		\includegraphics[width=\textwidth]{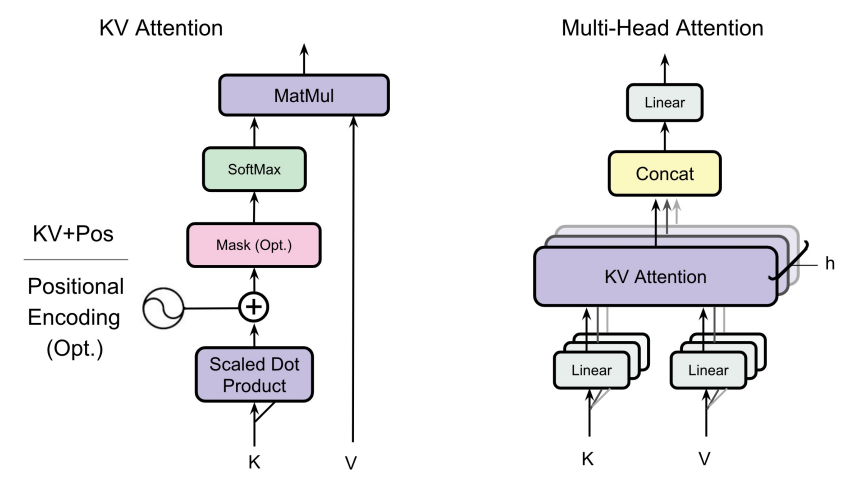}
		\caption{KV version of multi-head attention \cite{3628-borji_kv_transformer}.}
		\label{3628-fig:kv-att}
	\end{subfigure}
	\caption{We implemented an SETR encoder as well as a variant with KV multi-head attention.}

	\label{3628-fig:setr-and-kv-att}
\end{figure}

First introduced in 2017 by \citeauthor{3628-vaswani_attention_2017} \cite{3628-vaswani_attention_2017}, Transformers revolutionized the field of natural language processing (NLP) and subsequently computer vision through their novel use of self-attention mechanisms. \citeauthor{3628-dosovitskiy_16x16_words} \cite{3628-dosovitskiy_16x16_words} adapted the Transformer architecture to image classification tasks by splitting input images into fixed-size patches. Each patch is flattened into a vector and then linearly projected into a higher-dimensional representation using a linear transformation defined as: $Z = WP + b$ where $P$ is the flattened patch vector, $W$ is the weight matrix, $b$ is the bias vector, and $Z$ is the resulting high-dimensional representation.
To address the lack of spatial inductive bias inherent in Transformers compared to CNNs, positional embeddings are added to preserve the arrangement of the patches.
Once the patches are embedded, they are fed into a standard Transformer encoder, which applies self-attention mechanisms. The output of the self-attention layer can be computed using
\begin{align}
	\text{Attention}(Q, K, V) = \text{softmax}\left(\frac{QK^\top}{\sqrt{d_k}}\right)V
	\label{3628-eq:qkv-att}
\end{align}
where $Q$, $K$, and $V$ are the query, key, and value matrices respectively, and $d_k$ is the dimensionality of the keys.
Finally, the patch representations are processed through a multi-layer perceptron (MLP) head to convert the high-dimensional Transformer outputs back into task-specific predictions. For image classification, the prepended class token is processed by the MLP head to produce a single class prediction.
In 2020, \citeauthor{3628-zheng_SETR} \cite{3628-zheng_SETR} introduced SETR (SEgmentation TRansformer), with a purely transformer (ViT) based encoder for semantic segmentation (Fig. \ref{3628-fig:setr-encoder}). Semantic segmentation is achieved by use of a convolutional decoder with progressive upsampling (PUP). Herein, the high dimensional feature maps that are output from the transformer encoder are reconstructed into the dimension of the input image. The output of the encoder is a token sequence with a shape of $z_L \in \mathbb{R}^{N\times C}$ where $C$ is the feature dimension. This sequence is first reshaped into a feature map of shape $z_L \in \mathbb{R}^{\frac{H}{16} \times \frac{W}{16} \times C}$ and subsequently sent through four convolution layers, which each perform a $2 \times$ upsampling operation. The resulting feature map has the shape of the input image and a depth equal to the number of classes to be segmented ($H \times W \times C$) (Fig. \ref{3628-fig:setr-pup-adapted}).
In traditional Transformers, the self-attention mechanism computes three components, queries (Q), keys (K), and values (V), dynamically for every token in the sequence, thereby determining the importance of each token relative to others. \citeauthor{3628-borji_kv_transformer} \cite{3628-borji_kv_transformer} expands on the original transformer architecture and its various derivates by investigating whether the inclusion of queries, keys, and values are required in order to achieve competitive performance. The self attention mechanism in a standard transformer is given by Eq \ref{3628-eq:qkv-att}. The weight matrices $W_q, W_k, \text{and } W_v$ are the parameters responsible for projecting the input $X$ into an output tensor of dimension $d$, with $Q = XW_q, K = XW_k, \text{and } V = XW_v$. This formulation relies on computing the dot product between each token in the query ($Q$) and the corresponding token in the key ($K$). The approach proposed by \citeauthor{3628-borji_kv_transformer} \cite{3628-borji_kv_transformer} modifies this by replacing $Q$ with $K$, such that
\begin{align}
	\text{Attention}(K, V) = \text{softmax}\left(\frac{KK^\top}{\sqrt{d_k}}\right)V
	\label{3628-eq:kv-att}
\end{align}
holds true.
The KV self-attention mechanism initially creates a symmetric structure. To introduce asymmetry, \citeauthor{3628-borji_kv_transformer} \cite{3628-borji_kv_transformer} suggests augmenting the $n \times n$ attention matrix with 2D positional encoding of dimension $m$. This modification transforms the matrix into one with dimensions $n \times n \times m$. A linear layer, consisting of $m$ neurons, is then employed to project this expanded matrix back to its original $n \times n$ size (Fig. \ref{3628-fig:kv-att}).
Tab. \ref{3628-tab:complexity-kv-att} illustrates the expected advantages in complexity and computational cost associated with KV attention.
The KV+Pos attention results in higher parameter and FLOPS counts, which depend on the parameter $m$. It is important to note that $m$ can be set to a much lower value than $d$. However, a significant drawback of KV+Pos attention is that its computational complexity is dependent on $n^2$, which can lead to high computational costs. \cite{3628-borji_kv_transformer}
\begin{SCtable}
	\begin{tabular}{lll}
		\hline
		\multicolumn{1}{p{15mm}}{\multirow{2}{6em}{}} & \multicolumn{1}{c}{\multirow{2}{6em}{Comp. complexity}} & \multicolumn{1}{c}{\multirow{2}{6em}{\# Parameters}} \\
		                                              &                                                         &                                                      \\
		\hline
		QKV                                           & $2nd^2$                                                 & $2d^2$                                               \\
		KV                                            & $nd^2$                                                  & $d^2$                                                \\
		KV+pos                                        & $nd^2+n^2m$                                             & $d^2+m$                                              \\
		\hline
	\end{tabular}
	\caption{Comparison of computational complexity and number of parameters of attention mechanisms. Here, $d$ represents the embedding dimensionality, $n$ denotes the sequence length, and $m$ is the dimensionality of the 2D positional encoding layer. \cite{3628-borji_kv_transformer}}
	\label{3628-tab:complexity-kv-att}
\end{SCtable}
TransUNet \cite{3628-chen_TransUNet} proposes a hybrid (CNN and Transformer) model. The encoder leverages a ResNet-50 architecture \cite{3628-ResNet} to extract high-level features from input images.
Instead of splitting the image into patches,
TransUNet applies patch embedding to the feature maps generated by the CNN, creating feature patches as the transformer input. This approach allows for better preservation of spatial information and captures fine-grained details. The transformer layers then process these patches to model long-range dependencies, enhancing the network's ability to integrate global context with local features.
\citeauthor{3628-wu_cvt} \cite{3628-wu_cvt} introduce the Convolutional Vision Transformer (CvT), a hybrid model that enhances the Vision Transformer (ViT) by integrating two key convolutional operations: token embedding and projection.
In CvT, convolutional token embedding replaces ViT’s patch splitting by applying convolutional layers. This approach allows the model to effectively leverage local information by applying multiple filters that slide over the entire image.
The resulting feature maps maintain the spatial structure of the image while reducing its dimensionality.
Convolutional projection applies further convolutions to the feature maps prior to passing them to the self-attention mechanism.
This hybrid approach boosts accuracy, reduces parameter count, and enhances computational efficiency compared to the original ViT, particularly for image classification tasks.

\section{Materials and methods}
\subsection{Dataset}\label{3628-sec:dataset}
The medical image dataset used is UW-Madison GI Tract Image Segmentation \cite{3628-uw_madison_gi_tract} which consists of abdominal MRI slices. Annotations of the three classes (large bowel, small bowel, stomach) are provided in the form of run-length encoded organ segmentations. During preprocessing, we transformed the RLE ground truth data into 2D greyscale multi-class masks. Moreover, the dataset was split into training, validation, and test sets with a ratio of $80:16:4$.
\subsection{Model architectures}\label{3628-sec:model-arch}
\begin{figure}[hbtp]
	\centering
	\includegraphics[width=0.9\textwidth]{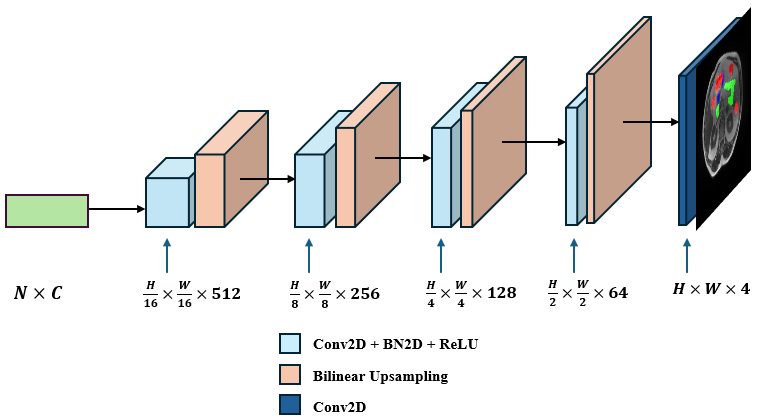}
	\caption{We adapted the SETR-PUP decoder to also reduce feature dimensions during upsampling.}
	\label{3628-fig:setr-pup-adapted}
\end{figure}

To explore KV attention across different architectures, we implemented several models with KV and QKV attention variants, as detailed below.
All models share a convolutional decoder adapted from SETR (Sec. \ref{3628-sec:related-work}).
We modified the decoder to halve the feature dimensions during upsampling, reducing the overall parameter count (Fig. \ref{3628-fig:setr-pup-adapted}).
We began by implementing the SETR encoder as outlined in \cite{3628-zheng_SETR}, using a ViT-B/16 backbone with the feature dimension $D = 768$, number of heads $H = 12$, and number of layers $L = 12$ with both QKV and KV attention mechanisms. We refer to these architectures as SETR-QKV and SETR-KV, respectively.
Furthermore, we explore SETR-KV-pos, where we introduce positional encoding within the KV attention block to create asymmetry, following the approach proposed in \cite{3628-borji_kv_transformer}. The 2D positional encoding dimension $m$ was set to 50.
Additionally, we constructed two models with a hybrid encoder. Drawing inspiration from TransUNet \cite{3628-chen_TransUNet}, we integrated the first four convolutional layers of the ResNet-50 architecture \cite{3628-ResNet} into our encoder to capture higher-dimensional features before the patch embedding stage. In the fourth layer, we increased the number of blocks from six to nine to improve feature extraction while maintaining a feature dimension of 1024. Unlike the approach in \cite{3628-chen_TransUNet}, no skip connections were used. We refer to these models as SETR-QKV-CE and SETR-KV-CE, respectively.
Finally, we developed an additional hybrid model using a Convolutional Vision Transformer (CvT) \cite{3628-wu_cvt} as the encoder. The models SETR-QKV-CVT and SETR-KV-CVT utilize a CvT-13 encoder, with the multi-head attention (MHA) in the Convolutional Transformer Blocks implemented with QKV and KV attention, respectively.
\subsection{Experiments}
All models described in Sec. \ref{3628-sec:model-arch} were trained for 100 Epochs without early stopping to ensure comparable results. The input resolution was set at $224\times224$ and a fixed patch size of $16\times16$ was chosen. The AdamW optimizer with a learning rate of $10^{-4}$ and polynomial learning rate scheduling with factor $0.9$ were used. Furthermore, a batch size of 32 was chosen for training. During training, on-the-fly data augmentation was applied, namely horizontal flipping, vertical flipping, shift scale rotate, coarse dropout, and random bright contrast, each having $50 \%$ probability of being applied.
All models were trained on the dataset described in Sec. \ref{3628-sec:dataset} without using any pretrained backbones.

\section{Results}
The performance metrics computed for the tested architectures include the Jaccard index and the weighted Jaccard index.
Model complexity is represented by the number of learnable parameters, while computational efficiency is assessed by the number of multiply-accumulate operations (MACs) (collected through the \verb|torchinfo| and \verb|ptflops| python modules). Our results indicate that all KV variants perform comparably well or slightly better than their corresponding QKV implementations, while also demonstrating a reduction in both parameter count and MACs of approximately $10\%$ (Tab. \ref{3628-tab:results}).
\begin{table}
	\caption{The results of our experiments. We observe no performance decrease among most of the KV variants, while simultaneously seeing a reduction in parameter count and MACs. The asterisk (*) indicates that the MACs calculation requires further revision, as it does not account for the calculations related to 2D positional encoding. }
	\label{3628-tab:results}
	\begin{tabular*}{\textwidth}{l@{\extracolsep\fill}cccc}
		\hline
		& Jaccard & Weighted Jaccard & \# Parameters (M) & MACs (G) \\
		\hline
		SETR-QKV      & 0.8718 & 0.8653 & 91.04   & 21.07 \\
		SETR-KV       & 0.8727 & 0.8667 & 83.96   & 19.68 \\
		SETR-KV-pos   & 0.8750 & 0.8691 & 83.96   & 19.97* \\
		SETR-QKV-CE   & 0.8990 & 0.8946 & 103.14  & 25.02 \\
		SETR-KV-CE    & 0.9015 & 0.8960 & 96.05   & 23.68 \\
		SETR-QKV-CVT  & 0.8846 & 0.8786 & 22.9    & 9.86 \\
		SETR-KV-CVT   & 0.8807 & 0.8755 & 21.3    & 9.49 \\
		\hline
	\end{tabular*}
\end{table}

\section{Discussion}
Our results indicate that KV Transformers demonstrate competitive performance on segmentation tasks compared to their QKV attention counterparts.
The incorporation of convolutional elements in the encoder leads to an anticipated increase in parameter count; however, this can be mitigated by utilizing KV attention.
The computational cost and complexity outlined in Tab. \ref{3628-tab:complexity-kv-att} were not realized in our experiments, as the inclusion of convolutional elements in the decoders increases both of these metrics.
Nonetheless, these findings motivate us to further investigate the application of KV attention in future projects.
Since the primary architectural change in KV attention takes place within the MHA block of the transformer, it could theoretically be adapted to various ViT-like architectures, presenting opportunities for additional research into the robustness of this design.
In conclusion, we have examined the use of KV Transformer architectures for segmentation tasks, particularly in the context of medical imaging.
Our findings demonstrate that the evaluated KV variants exhibit similar performance compared to their QKV counterparts, while also achieving reductions in both parameter count and computational cost.

\printbibliography

\end{document}